# Recognition of Smoking Gesture Using Smart Watch Technology

Casey A. Cole[1], Bethany Janos[2], Dien Anshari[3], James F. Thrasher[3], Scott Strayer[4], and Homayoun Valafar[1*]

[1]Department of Computer Science and Engineering, University of South Carolina, Columbia, SC 29208, USA

[2]Department of Biomedical Engineering, University of South Carolina, Columbia, SC 29208, USA

[3]Department of Public Health, University of South Carolina, Columbia, SC 29208, USA

[4]University of South Carolina School of Medicine, Columbia, SC 29208, USA

[*] Corresponding Author Email: homayoun@cec.sc.edu Phone: 1 803 777 2404 Fax: 1 803 777 3767

Mailing Address: Swearingen Engineering Center, Department of Computer Science and Engineering, University of South Carolina, Columbia, SC 29208, USA

**Abstract** – *Diseases resulting from prolonged smoking are the most common preventable causes of death in the world today. In this report we investigate the success of utilizing accelerometer sensors in smart watches to identify smoking gestures. Early identification of smoking gestures can help to initiate the appropriate intervention method and prevent relapses in smoking. Our experiments indicate 85%-95% success rates in identification of smoking gesture among other similar gestures using Artificial Neural Networks (ANNs). Our investigations concluded that information obtained from the x-dimension of accelerometers is the best means of identifying the smoking gesture, while y and z dimensions are helpful in eliminating other gestures such as: eating, drinking, and scratch of nose. We utilized sensor data from the Apple Watch during the training of the ANN. Using sensor data from another participant collected on Pebble Steel, we obtained a smoking identification accuracy of greater than 90% when using an ANN trained on data previously collected from the Apple Watch. Finally, we have demonstrated the possibility of using smart watches to perform continuous monitoring of daily activities.*

## 1 Introduction

In the past decade, measures have been taken to warn the population about the dangers of smoking. While the smoking rate has decreased significantly since then, smoking remains the leading preventable cause of death throughout the world. Additionally, youth tobacco use has increased as the popularity of products such as e-cigarettes and hookah has risen[1] . In America, 53.4% of college students have smoked at least one cigarette and 38.1% reported smoking in the past year[2] . Even though the hazards of smoking are generally accepted, there remains many smokers who struggle to quit. Those who try to quit are typically middle aged and beginning to feel the adverse effects of smoking. Yet, on average, smokers relapse four times before successfully quitting[3] . Many smokers do not realize that it is normal to require multiple attempts to quit smoking and therefore need recurring intervention and support to aid them. Constant support from an individual's community is shown to increase the likelihood of quitting[2] . The existence of an application (housed on a smart phone or watch) that would provide this constant support could greatly increase a person's fortitude to abstain from smoking.

The first step in making such an application is the ability to detect when a person is smoking so that the appropriate intervention can be initiated. Previous works have shown the possibility of detecting smoking gestures using in-house designed wearable devices[4,5]. These techniques have shown great promise with both high accuracy (95.7-96.9%) and low false positive rates (<1.5%). However, they require the use of devices not commonly found in a typical household such as multiple 9-axis inertial measurement units (IMU's), respiration bands that must be worn across the chest and two-lead electrocardiograph worn under the clothes. The use of these uncommonly and relatively expensive devices severely limits mass deployment for daily use.

Smart watches are becoming increasingly prevalent in common households. According to Apple's website, over 5 million Apple Watches were sold in 2015 alone and projections into 2016 show promising growth. Other smart watch companies like Asus and Pebble have seen similar growth patterns from as well. By contrast to the previous methods, the method explored in this study relies solely on the use of a smart watch's built-in accelerometer, effectively eliminating the need to use more uncommon detection devices. In addition, the pairing of a smartwatch with a smart mobile device enables immediate alerting, engagement and recruitment of social support groups to prevent or alter one's smoking behavior. As the first step in this process we have examined the feasibility and complexity of detection of smoking gesture using smart wearable devices. Our investigation has included minimum data requirement and an exploration of most informative dimension of accelerometer sensors. Prior knowledge of the problem complexity will allow for a smoother transition into actual deployment of our detection mechanism on smart watches in the future.

## 2 Background and Method

The overall view of our study consisted of three major stages: data collection, training of multiple artificial neural networks for pattern recognition, and evaluation. The following sections provide a more detailed outline for each of these stages.

### 2.1 Data Collection

Data in this study were acquired by a non-smoking participant utilizing an Apple Watch (version 2.1). Using the application PowerSense (available in App Store for iOS) a number of individual smoking gestures (also referred as a puff) and continuous smoking sessions (a session that consists of multiple puffs) were recorded and analyzed. All samples were measured at a 50Hz sampling rate. Due to minor fluctuations in the duration of each gesture, the number of data points varied for each gesture. Each isolated puff pattern was represented by 200 interpolated data point in order to create a uniform size input set for the pattern recognition stage. The resulting smoking gestures are shown below in Figure 1. The three differently colored line clusters represents each of the three dimensions of the accelerometer (X in blue, Y in red and Z in green). Each cluster is an overlay of all 20 individual smoking sessions used in the training of the neural networks. Based on visual inspection, it is clear that the smoking pattern is very well conserved across each of the samples.

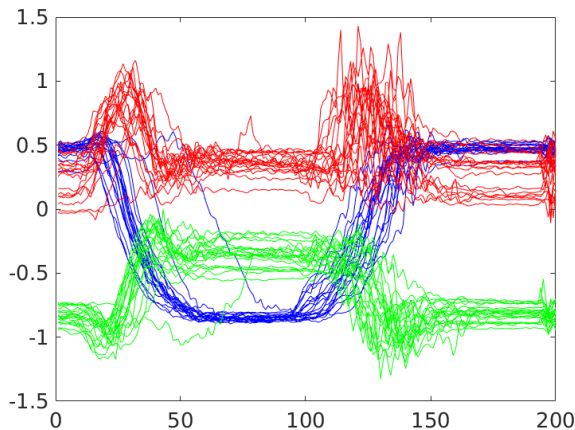

Figure 1. Overlay of all single smoking gestures with the X dimension in blue, Y in red and Z in green.

In addition to smoking sessions, several non-smoking gestures were also recorded. These gestures (seen in Figure 2) included drinking, scratching one's nose, yawning, coughing, brushing hair behind one's ear and rubbing one's stomach. The selection of these patterns were based on activities that may be similar to smoking gesture, or ones that may be present during most common smoking sessions. These gestures were including in both the training test and testing set.

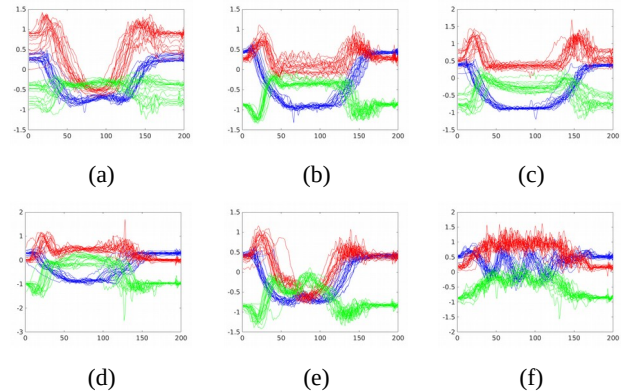

Figure 2. Overlay of all patterns collected for the following non-smoking gestures: (a) drinking, (b) scratching one's nose, (c) yawning, (d) coughing, (e) brushing hair behind one's ear, and (f) rubbing one's stomach.

In some of these gestures, such as scratching one's nose and yawning in Figure 2(b) and Figure 2(c) respectively, the movement of the hand and arm clearly resemble an individual smoking gesture (seen in Figure 1). Inclusion of these gestures into the data set will allow for studying how well the proposed method can distinguish smoking gestures from other very similar gestures.

In addition to individual gestures, longer continuous sessions were recorded. The continuous smoking sessions consisted of approximately 7 to 10 gestures per session. The non-smoking sessions included three common activities: eating, drinking and putting on chapstick/lipstick. Each session was divided into 200 time step segments using a rolling window approach. As seen in Figure 3, a continuous smoking session is no more than a combination of individual smoking gestures seen in Figure 1.

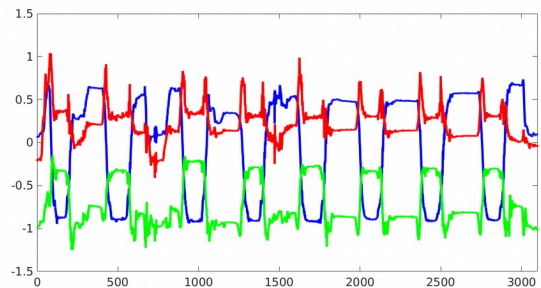

Figure 3. Example of continuous smoking session.

Table 1 summarizes the total number of smoking and non-smoking gestures in each of the data sets and breaks down the exact amount used in each phase of the investigation.

Table 1. Summary of data utilized in analyses.

|  | Non-smoking | Smoking |
|---|---|---|
| Training Set | 120 | 20 |
| Testing Set | 30 | 10 |
| Extended Sessions | 5 | 5 |

In order to test the applicability of the presented method across other wearable platforms, as well as other participants, one smoking session was acquired by a different participant (than the original set of data used for training) on the Pebble Time Steel used on Android platform. The Pebble Time Steel was selected as a second test watch due to its long lasting battery life, durability, and reasonable price. A continuous smoking session was recorded using the AccelTool (http://mgabor.hu/accel/) App at a sampling rate of 50 Hz.

## 2.2 Pattern Recognition Via Artificial Neural Networks

Neural networks have been shown to be an effective classifier in many biological and health informatics applications.[6–10] The neural network toolbox in Matlab (version R2016a) was utilized during this phase of our study. Levenberg-Marquardt backpropagation[11–13] was selected as the training algorithm in all sessions. For each training session, the data were randomly partitioned into 3 sets: 70% in the training set, 15% in the validation set and 15% in the testing set. The networks were then trained, validated and then rigorously tested for accuracy. The procedures for the training and validation/testing phases are outlined below.

*Training* – The interpolated raw data consisted of information from three dimensions (X, Y and Z). In order to fully identify and evaluate useful information in the data, all three dimensions were utilized both individually and in combination with each other. In total, five ANNs were created for use in this study—one for each of the three dimensions (X, Y and Z), one for the combination of all three dimensions (referred to as XYZ) and one for the average of the three dimensional data (referred to as AVG). The number of inputs for the X, Y, Z and AVG data sets was 200, while the input size for the XYZ data set was 600. In each of the neural networks the hidden layer consisted of 10 hidden neurons. A single output neuron was used, with zero denoting a non-smoking gesture and one signifying a smoking gesture.

*Validation/Testing* – Validation of the appropriate level of training was accomplished using 15% of the excluded training dataset. The networks were further subjected to testing using several different data sets. The first of which was the remaining 15% of the data excluded from the original training set. Next, a new set of individual gestures (not included in the original training set) was presented to the networks. To test the method on more realistic cases, continuous smoking and non-smoking sessions were also presented to the networks. Lastly, a continuous smoking session from a different smart watch was tested on each of the ANNs. The results of each test set are reported in Section 3.

## 2.3 Evaluation

To measure the success of the proposed method specificity, sensitivity and total accuracy of each trial were observed. Specificity describes the rate at which the method is able to correctly classify a non-smoking event. Specificity is calculated by use of Eq (1), where *TN* and *FP* denote the number of true negatives and false positives, respectively.

$$Specificity = \frac{TN}{TN+FP} \quad (1)$$

Sensitivity refers to the rate at which the method correctly identifies a smoking event and can be calculated using Eq (2), where *TP* represents the number of true positives and *FN* denotes the number of false negatives.

$$Sensitivity = \frac{TP}{TP+FN} \quad (2)$$

Total Accuracy is then a measure of how often the method correctly classifies both smoking and non-smoking gestures and is calculated by Eq (3).

$$Accuracy = \frac{TP+TN}{TP+TN+FP+FN} \quad (3)$$

In this work a cutoff threshold of 0.8 (or 80%) was used above which was considered a successful detection.

## 3 Results and Discussion

In the following sections the results of testing the neural networks are reported. In each section the data set that performed the best and worst are discussed.

### 3.1 Accuracy on Training Set

For each of the data sets the corresponding neural networks was independently trained 10 times and the network with the highest accuracy was chosen to be used for future test sets. The accuracies, specificities and sensitivities described in Figure 4 represent the performance of the final trained neural networks on their respective training sets.

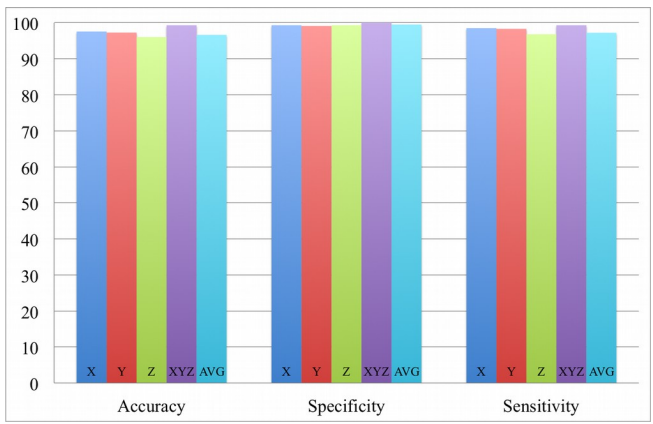

Figure 4. Accuracy, specificity and selectivity of the neural networks during training. The bars are individually labeled based on their respective training sets.

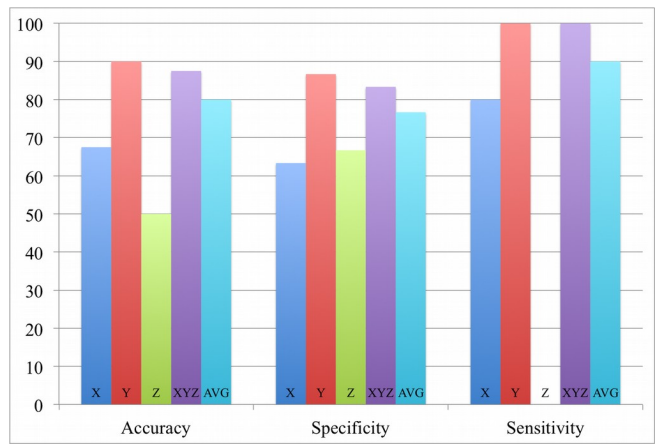

Figure 5. Accuracies, specificity and sensitivity in the individual gesture detection trials.

All of the neural networks exhibited high accuracy and specificity (> 90%) with respect to the training set of data. In each case the network trained with the XYZ data performed the best and the network trained with just the Z data performed the worst. A visual comparison of the individual smoking gestures and the non-smoking gestures clearly explains Z's poorer performance in correctly classifying smoking gestures. The Z dimension (in green Figures 1 and 2) exhibits very similar patterns across both smoking and non-smoking gestures. However, it is worth noting that the Z dimension still obtained a high specificity (nearly 100%) which means that it still may carry some complementary information, especially when identifying a non-smoking gesture.

## 3.2 Individual Gesture Detection

In this experiment, a new set of individual gestures (smoking and non-smoking) were presented to the previously trained neural networks. The accuracy, specificity and sensitivity are reported in Figure 5. Accuracies were measured by forward propagating each of the new samples in the corresponding neural network and then recording the number of correct and incorrect predictions. A threshold of 0.5 was used in interpretation of the neural network output, that is, any output larger than 0.5 was considered as smoking and any output lower than 0.5 was considered as non-smoking.

As shown in Figure 5, the Y dimension performed the best with not only the highest accuracy, but also the highest specificity and a 100% sensitivity. The XYZ and AVG data sets also performed well especially in their ability to identify smoking gestures. Consistent with the previous section, the Z dimension performed the worst with both a low accuracy and specificity as well as a 0% sensitivity rate indicating utilization of the Z dimension results in identification of 0/10 smoking gestures.

To better understand the nature of false-positive classifications, contribution of each individual gesture was recorded and results are shown in Figure 6. Based on the results shown in this figure, the non-smoking pattern that caused the most false positives was coughing followed by scratching of nose and yawning. These results were expected due to the high degree of visual similarity of these gestures to an individual smoking gesture.

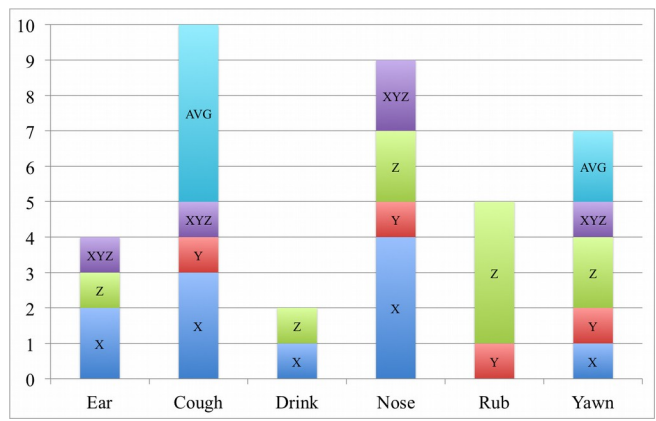

Figure 6. Total number of false positives created by each non-smoking gesture. Each segment is labeled based on the dimension of the accelerometer data being used.

## 3.3 Continuous Gesture Detection

In this section the results for detection on continuous monitoring of gestures are reported. To accomplish this objective the neural networks trained on static gestures were utilized. Using a running window of size 200 (without any interpolation) the continuous gestures were parsed into data sets for input into the networks. The classification result for each running window (0 denoting non-smoking and 1 denoting smoking) is plotted at the beginning of each running window. Figure 7 illustrates an example output (in purple before converting to a binary representation) of the neural network trained on X. Visual inspection of this figure clearly confirms the correlation between spikes in detection pattern over the regions where an apparent smoking gesture. However, there is not a trivial way to quantify the network's success because it is not immediately clear where should constitute the start and end of a gesture. Therefore, in order to be sure to encompass the entire gesture, generous ranges were handpicked to describe each smoking gesture. As in the previous section, a cutoff of 0.5 was chosen where any output greater than 0.5 was considered smoking and anything below was counted as non-smoking. Specificity was measured by considering all other sections of the continuous gesture not within the smoking ranges.

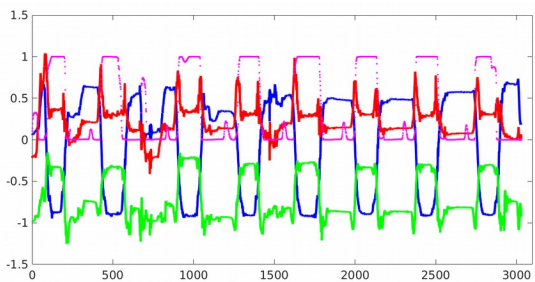

Figure 7. Example of continuous smoking session superimposed with the output of the neural network trained on the X dimension.

The averaged results across all five continuous sessions are presented in Figure 8 along with error bars representing the minimum and maximum of each averaged result.

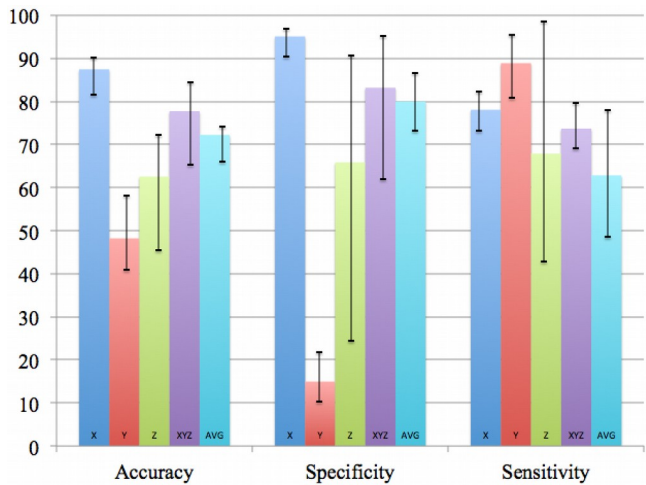

Figure 8. Averaged accuracies, specificities and sensitivities across the five continuous smoking gesture detection experiments with error bars representing the respective min and max of each value.

In the continuous smoking sessions, the X dimension performed better than all other dimensions. The Y dimension had a better sensitivity score but this can be disregarded due to its low specificity score. A high sensitivity coupled with a low specificity score denotes that the Y dimension classifies practically everything as smoking and therefore it's high sensitivity should be ignored. In this sense, Y performed the worst overall.

In the non-smoking sessions selectivity becomes inapplicable and specificity is equivalent to total accuracy. therefore, only accuracies these sessions are presented in Figure 9.

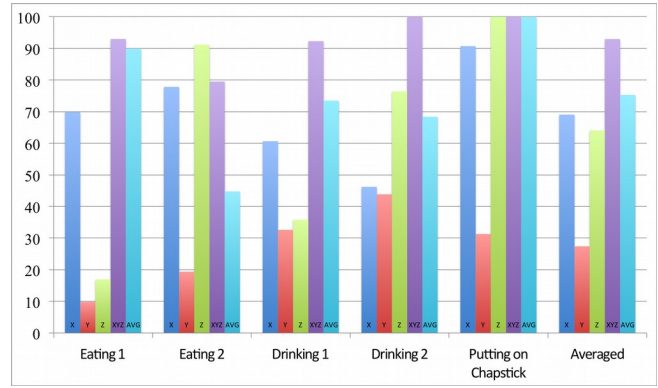

Figure 9. Accuracies for five continuous non-smoking session trials.

Despite it's superior performance in classification with continuous smoking gestures, the X dimension performed with an average accuracy of under 70% in the non-smoking continuous sessions. It seems that in the presence of more complex continuous gestures (like eating and drinking) the XYZ data set seems to contain the most useful information for correct classification. Eating sessions seemed to pose the most difficulty for XYZ. This could signify that eating is one of the closest gestures to smoking and can therefore lead to confusion in the network. In accordance with previous results, the Y dimension performed the worst across all the non-smoking sessions.

## 3.4 Exploration of dependency on the wearable device

As described in the Section 2.1, a continuous smoking gesture was recorded using a Pebble smart watch. Results were collected using the pre-existing neural networks that was trained on data from the Apple Watch from a different participant. Figure 10 shows the smoking session recorded using the Pebble watch, which exhibit significant similarity to patterns shown in Figure 1. The outputs of the neural network using the X data set are shown in purple. Again, there is good correlation between the smoking gestures and the spikes in the output of the neural network. Figure 11 shows the resulting accuracies, specificities and sensitivities for this smoking session.

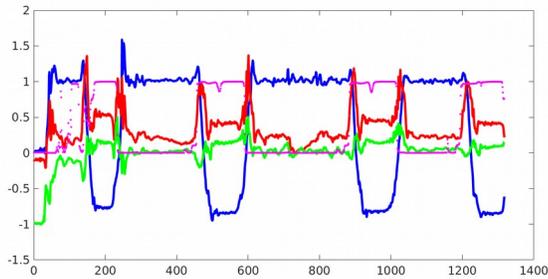

Figure 10. Output of the neural network for the X dimension superimposed to the original smoking session.

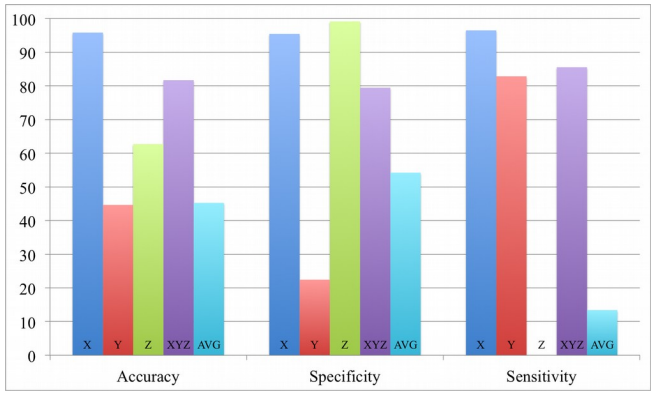

Figure 11. Results for the Pebble smart watch.

As in the previous cases of continuous smoking sessions, the X dimension performed the best in classifying the gestures. In the case of the Pebble watch, the Y and Z dimensions did equally poorly. The Y dimension classified almost everything as smoking and the Z dimension classified nearly everything as non-smoking.

## 4    Conclusion

The general summary of our work supports the feasibility in detection of smoking gestures using typical sensors available in smart watches. Based on our experiments, pattern recognition via Artificial Neural Networks applied to the sensor data obtained from smart watches can produce performances very comparable to previously reported work. However, the use of a smart watch is far more pragmatic in general population studies over the other existing technologies. Results shown in section 3.4 suggest the possibility of delivering an application capable of detecting smoking gesture across the general population of smokers. This universal Artificial Neural Network eliminates the need to customize a training session per user.

Our exploration of efficacy of individual sensor data in detection of gesture has produced unexpected results. The neural network trained with data from just the X dimension performed the best in the presence of continuous smoking gestures but when faced with more complex non-smoking motions, it fails and a more complete set of data is needed to distinguish smoking gestures. Across all the testing sets the neural network trained with data from all three dimensions (XYZ) did consistently well. However, the XYZ data set requires 600 inputs to the network whereas the X data set only requires 200. This is a significant reduction in data requirement which directly impacts the computational time of the method. For this reason, the viability of both data sets will continue to be investigated.

Additional investigations are required before general deployment of such approaches. Continuous monitoring of data may be outside of power limitations of such devices and may act as a technological barrier. Our future investigations will include optimization of sampling rate, minimization of bluetooth communication between the smart watch and the companion phone, and better assessment of the universality of the trained ANN.

## 5    Acknowledgments

Funding for this work is provided by ASPIRE-II grant from the University of South Carolina Research Foundation.

## 6    Bibliography

1.  Li, K. *et al.* Smoking and Risk of All-cause Deaths in Younger and Older Adults: A Population-based Prospective Cohort Study Among Beijing Adults in China. *Medicine (Baltimore).* **95**, e2438 (2016).

2.  Rooney, B. L., Silha, P., Gloyd, J. & Kreutz, R. Quit and Win smoking cessation contest for Wisconsin college students. *WMJ* **104**, 45–9 (2005).

3.  Khati, I. *et al.* What distinguishes successful from unsuccessful tobacco smoking cessation? Data from a study of young adults (TEMPO). *Prev. Med. reports* **2**, 679–85 (2015).

4.  Saleheen, N. *et al.* puffMarker. in *Proceedings of the 2015 ACM International Joint Conference on Pervasive and Ubiquitous Computing - UbiComp '15* 999–1010 (ACM Press, 2015).

5.  Parate, A., Chiu, M.-C., Chadowitz, C., Ganesan, D. & Kalogerakis, E. RisQ: Recognizing Smoking Gestures with Inertial Sensors on a Wristband. *MobiSys ... ... Int. Conf. Mob. Syst. Appl. Serv. Int. Conf. Mob. Syst. Appl. Serv.* **2014**, 149–161 (2014).

6.  Valafar, H. *et al.* Predicting the effectiveness of hydroxyurea in individual sickle cell anemia patients. *Artif. Intell. Med.* **18**, 133–48 (2000).

7.  Fawcett, T. M., Irausquin, S. J., Simin, M. & Valafar, H. An artificial neural network approach to

improving the correlation between protein energetics and the backbone structure. *Proteomics* **13**, 230–8 (2013).

8. Fawcett, T. M., Irausquin, S., Simin, M. & Valafar, H. An Artificial Neural Network Based Approach for Identification of Native Protein Structures Using an Extended ForceField. *Proc. 2011 IEEE Int. Conf. Bioinforma. Biomed. (Atlanta, Georg. USA, Novemb. 12-15, 2011)* (2011).

9. Valafar, F., Valafar, H., Ersoy, O. K. & Schwartz, R. G. Comparative studies of two neural network architectures for modeling of human speech production. in *Proceedings of ICNN'95 - International Conference on Neural Networks* vol. 4 2056–2061 (IEEE).

10. Valafar, H., Arabnia, H. R. & Williams, G. Distributed global optimization and its development on the multiring network. *Neural, Parallel Sci. Comput.* **12**, 465–495 (2004).

11. Levenberg, K. A method for the solution of certain problems in least squares. *Q. Appl. Math.* **2**, 164–168 (1944).

12. Marquardt, D. W. An Algorithm for Least-Squares Estimation of Nonlinear Parameters. *J. Soc. Ind. Appl. Math.* **11**, 431–441 (1963).

13. Ranganathan, A. The Levenberg-Marquardt Algorithm. (2004).